\documentclass[runningheads]{llncs}

\usepackage{eccv}
\usepackage{eccvabbrv}

\usepackage{graphicx}
\usepackage{booktabs}

\usepackage[accsupp]{axessibility}
\usepackage{picinpar}
\usepackage{multirow}
\usepackage{colortbl}

\usepackage[pagebackref,breaklinks,colorlinks,citecolor=eccvblue]{hyperref}
\usepackage{orcidlink}
\usepackage{marvosym}

\newcommand{\rblue}{\rowcolor{blue!10}}
\newcommand{\name}{CLTD}

\begin{document}

\title{Causality-inspired Discriminative Feature Learning in Triple Domains for Gait Recognition} 

\titlerunning{CLTD}

\author{Haijun Xiong\orcidlink{0000-0003-0856-8250} \and
Bin Feng\textsuperscript{(\Letter)}\orcidlink{0000-0003-2166-751X} \and
Xinggang Wang\orcidlink{0000-0001-6732-7823} \and
Wenyu Liu\orcidlink{0000-0002-4582-7488}}

\authorrunning{H.~Xiong et al.}
\institute{Hubei Key Lab of Smart Internet, School of EIC, Huazhong University of Science and Technology, Wuhan, China \\
\email{\{xionghj,fengbin,xgwang,liuwy\}@hust.edu.cn}}

\maketitle

\begin{abstract}
Gait recognition is a biometric technology that distinguishes individuals by their walking patterns. However, previous methods face challenges when accurately extracting identity features because they often become entangled with non-identity clues. To address this challenge, we propose \name{}, a causality-inspired discriminative feature learning module designed to effectively eliminate the influence of confounders in triple domains \ie, spatial, temporal, and spectral. Specifically, we utilize the Cross Pixel-wise Attention Generator (CPAG) to generate attention distributions for factual and counterfactual features in spatial and temporal domains. Then, we introduce the Fourier Projection Head (FPH) to project spatial features into the spectral space, which preserves essential information while reducing computational costs. Additionally, we employ an optimization method with contrastive learning to enforce semantic consistency constraints across sequences from the same subject. Our approach has demonstrated significant performance improvements on challenging datasets, proving its effectiveness. Moreover, it can be seamlessly integrated into existing gait recognition methods.
\end{abstract}

\section{Introduction}
\label{sec: intro}
Gait recognition is a long-distance biometric identification technology that uses unique walking patterns for individual recognition. It has garnered significant attention due to its wide applications in surveillance and healthcare~\cite{shen2022comprehensive,sarkar2005humanid}. However, accurately identifying individuals presents a challenge due to various factors influencing performance.

\begin{figure}[tb]
  \centering
  \includegraphics[width=0.75\linewidth]{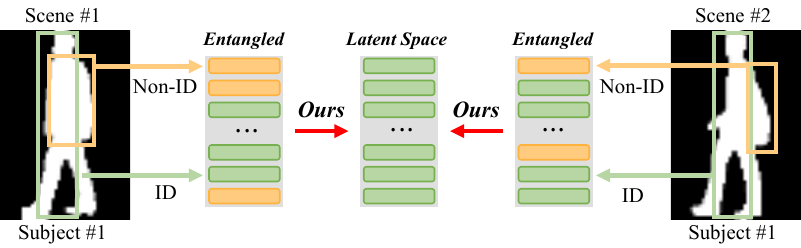}
  \caption{Motivation. Illustration of entanglement between non-ID and ID clues. With our approach, the impact of non-ID clues is systematically eliminated.}
  \label{fig: ID and non-ID}
\end{figure}

There are two categories of gait recognition methods: appearance-based~\cite{chao2019gaitset,huang2021context,fan2023opengait,ma2023dynamic} and model-based~\cite{guo2023physics,teepe2021gaitgraph,fu2023gpgait,huang2023condition}. Appearance-based approaches have gained more attention recently because of their superior performance compared to model-based approaches. However, the entanglement of discriminative identity (ID) features with non-identity (non-ID) clues makes it difficult to effectively and efficiently extract features. Here, non-ID clues such as noise, clothing, and carrying conditions, act as extrinsic confounders to identities. As depicted in~\cref{fig: ID and non-ID}, most of the current approaches focus on extracting ID-intrinsic features directly from the entangled space, which is challenging. This entanglement biases gait recognition models towards non-ID clues, ultimately compromising performance~\cite{li2020gait}. The core idea of this work is to establish a consistent latent space where non-ID clues can be effectively eliminated.

The Structural Causal Model (SCM) provides a fundamental framework for describing causal mechanisms within a system~\cite{zhang1994probabilistic}, with applications spanning diverse fields such as computer vision~\cite{yang2023good}, natural language processing~\cite{abraham2022cebab}, and recommender systems~\cite{ding2022causal}. Recently, causal analysis has been explored in gait recognition, with notable contributions from the Generative Counterfactual Intervention framework (GaitGCI)~\cite{dou2023gaitgci}. GaitGCI focuses on mitigating the influence of confounders during the feature learning process, but it has several limitations. Firstly, it only addresses confounders at the output stage of the backbone network, overlooking the intricate formation and effects of confounders across different stages. Secondly, its causal intervention is confined to spatial features, neglecting potential confounders in other domains such as temporal and spectral. Lastly, the absence of explicit semantic consistency constraints may compromise the consistency of learned identity representations for the same subject, hindering overall performance.

In contrast, we propose to leverage the relationship between spatial and spectral domain signals. We believe that if confounders exist in spatial signals, they are also likely present in the spectral domain. We can effectively describe and separate signals by exploiting spectral representation, thereby eliminating confounders. This notion has been validated in various computer vision tasks~\cite{UHDFourICLR2023,lee2023decompose,zhou2023fourmer,mao2023intriguing}, underscoring its potential efficacy in gait recognition.

Based on these analyses, we propose a novel module called \name{}, which is designed to mitigate the influence of confounders within spatial, temporal, and spectral domains. First, we perform the Cross Pixel-wise Attention Generator (CPAG) to capture contextual information for each pixel in the time, height, and width dimensions, thereby generating high-quality attention distributions. This ensures the effective processing of spatial and temporal information. Next, we present the Fourier Projection Head (FPH), which utilizes the Fast Fourier Transform (FFT) to project the spatial feature map into the Fourier spectral domain. This transformation preserves essential information while significantly reducing computational costs, enabling efficient processing in the spectral domain. Furthermore, we introduce contrastive learning to supervise the generated factual and counterfactual features, thus ensuring consistent semantic representations of ID-intrinsic clues across sequences from the same subject. Moreover, we propose to utilize \name{} at multiple stages of the network, not just in the final stage, to comprehensively eliminate the effects of confounders. Importantly, \name{} serves as a plug-and-play training paradigm, compatible with various gait recognition methods. We validate the effectiveness and versatility of \name{} through experiments conducted on multiple datasets, including OU-MVLP~\cite{takemura2018multi}, CASIA-B~\cite{yu2006framework}, GREW~\cite{zhu2021gait}, and Gait3D~\cite{zheng2022gait}. The results demonstrate significant performance improvements when \name{} is integrated with various gait recognition methods, leading to state-of-the-art results on multiple datasets, with a maximum improvement of \textbf{11.1\%} (from 60.2\% to 71.3\%, refer to~\cref{tab: versatility}).

The main contributions of our work can be summarized as follows:
\begin{itemize}
    \item We solve the gait recognition problem from a new perspective, \ie, causality-based contrastive learning. This way aims to establish a consistent latent space, reducing the bias introduced by non-ID clues.
    \item We propose \name{}, which leverages the Cross Pixel-wise Attention Generator and Fourier Projection Head to eliminate the impact of confounders in triple domains effectively.
    In addition, we incorporate contrastive learning to ensure the consistency of ID-intrinsic semantics information across sequences from the same subject.
    \item Demonstrating the effectiveness and versatility of \name{} through extensive experiments, achieving state-of-the-art results on various datasets and significantly improving performance on multiple baseline methods.
\end{itemize}

\section{Related Work}
\label{sec: related}
\noindent \textbf{Gait Recognition.} Gait recognition methods can be generally categorized into two groups: model-based and appearance-based. Model-based approaches~\cite{guo2023physics,teepe2021gaitgraph,teepe2022towards,fu2023gpgait,huang2023condition,zhang2023spatial} use a priori model representing human shape and dynamics, extracting gait features by fitting these models. For instance, methods in~\cite{teepe2021gaitgraph,teepe2022towards} utilize ResGCN for spatial-temporal feature extraction from skeleton sequences. In contrast, GPGait~\cite{fu2023gpgait} focuses on efficient fine-grained feature extraction. On the other hand, appearance-based methods~\cite{han2005individual,shiraga2016geinet,chao2019gaitset,fan2020gaitpart,lin2021gait,wang2023hierarchical,zheng2022gait,huang2022star,dou2023gaitgci,huang2021context,chai2022lagrange,ma2023dynamic,hou2020gait} directly extract gait features from silhouette sequences. Han \etal~\cite{han2005individual} introduced Gait Energy Images (GEIs) obtained by averaging multiple silhouettes from a gait cycle sequence. GaitSet\cite{chao2019gaitset} is the first method to treat gait as a set for learning discriminative feature representations. Huang \etal~\cite{huang2021context,huang2022star} presented fine-grained feature representations and multi-scale temporal motion modeling methods. SMPLGait~\cite{zheng2022gait} introduces a multimodal method, which uses 3D human meshes~\cite{loper2015smpl} to enhance silhouette information. Chai \etal~\cite{chai2022lagrange} implemented a second-order motion extraction module and view embedding based on 3D convolution to learn motion information. Wang \etal~\cite{wang2023dygait} presented DyGait focusing on extracting feature representations of dynamic body parts. Additionally, some other methods~\cite{wang2023hierarchical,lin2021gait} explored the influence of part-based multi-scale motion information.

\noindent \textbf{Causality in Computer Vision.} Causality has emerged as a crucial concept in various computer vision tasks, aiding in the representation and reasoning of uncertain knowledge. Several recent works have leveraged causal models to address different challenges. Yang \etal~\cite{yang2021causal} proposed a causal attention model to eliminate confounding effects in vision-language attention models, enhancing the interpretability and robustness of the model. Chen \etal~\cite{chen2023meta} presented a meta-causal learning method that learns to infer the causes of domain shift between the auxiliary and source domains during training, facilitating domain adaptation and transfer learning tasks. Miao \etal~\cite{miao2023caussl} developed CauSSL a framework for semi-supervised medical image segmentation, which enhances algorithmic independence between two branches using causal mechanisms. Dou \etal~\cite{dou2023gaitgci} proposed GaitGCI, first introducing causality into gait recognition. GaitGCI focuses on intervening causally on the final spatial feature to mitigate confounding effects. In contrast to GaitGCI, our proposed approach eliminates the impact of confounders in triple domains, namely spatial, temporal, and spectral domains. This allows for a more comprehensive elimination of confounders and improves the discriminative ability of the gait recognition model.

\noindent \textbf{Contrastive Learning.} Contrastive learning is a powerful technique, aiming to decrease the distance between similar samples while increasing the gap between dissimilar ones. Guo \etal~\cite{guo2022contrastive} proposed AimCLR, emphasizing extreme augmentations to improve the universality of learned representations. Rao and Miao~\cite{rao2023transg} presented TranSG to capture skeleton relationships and spatial-temporal semantic information. Quan \etal~\cite{quan2023semantic} introduced SemCL, specifically targeting object extraction, significantly improving the spatial comprehension of pre-trained models. Inspired by these, \name{} ensures semantic consistency between sequences of the same subject by incorporating contrastive learning.

\section{Method}
\label{sec: method}
\subsection{Gait Recognition from Causal View}

\begin{figwindow}[0,r,{
    \includegraphics[width=0.35\linewidth]{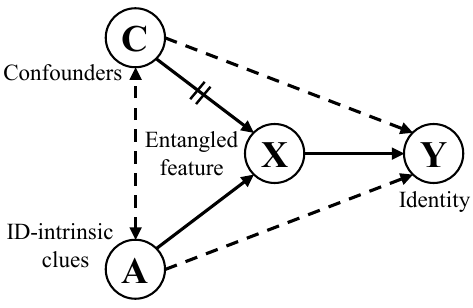}},
    {Causal Graph illustration of our approach.}
    \label{fig: Causal Graph}]
    \noindent Causal inference~\cite{pearl2016causal} refers to the process of determining the cause-and-effect relationship between different factors or events. Enlightened by previous excellent works~\cite{yang2023good,chen2023meta,li2023causally,lee2023mitigating}, we introduce causality into gait recognition to separate various confounders and ID-intrinsic clues. The causal graph in~\cref{fig: Causal Graph} is constructed using SCM~\cite{zhang1994probabilistic} to understand causality in gait recognition. In this representation, $\boldsymbol{X}$ denotes features generated by the gait model, $\boldsymbol{A}$ denotes ID-intrinsic clues, $\boldsymbol{C}$ stands for confounders (non-ID clues) entangled with $\boldsymbol{A}$, and $\boldsymbol{Y}$ represents the ground truth, \ie, the identity. The solid line indicates a direct causal relationship, \eg, $\boldsymbol{X} \to \boldsymbol{Y}$ denotes $\boldsymbol{Y}$ is caused by $\boldsymbol{X}$, while the dotted line indicates there exists statistical dependence between two factors. In the ideal scenario, $\boldsymbol{X}$ should be learned solely from ID-intrinsic clues, denoted as $\boldsymbol{A} \to \boldsymbol{X} \rightarrow \boldsymbol{Y}$. However, in practical cases, the existence of confounders often entangles $\boldsymbol{A}$ with non-ID clues $\boldsymbol{C}$, represented as $(\boldsymbol{C}, \boldsymbol{A}) \to \boldsymbol{X} \to \boldsymbol{Y}$, where $(\boldsymbol{C}, \boldsymbol{A})$ denotes that the two factors are entangled. It is difficult to distill $\boldsymbol{X}$ from confounders $\boldsymbol{C}$ due to the severe entanglement of various influencing factors with $\boldsymbol{A}$. Performing intervention on confounders $\boldsymbol{C}$, such as \textit{do}$\sim$operation denoted as $do(\boldsymbol{C})$ for specifying the exact value of $\boldsymbol{C}$ and isolating its cause, allows us to eliminate the corresponding effects. In~\cref{fig: Causal Graph}, $\vert\vert$ denotes the \textit{do}$\sim$operation.
\end{figwindow}

\indent

\begin{figure}[tb]
    \centering
    \includegraphics[width=0.85\linewidth]{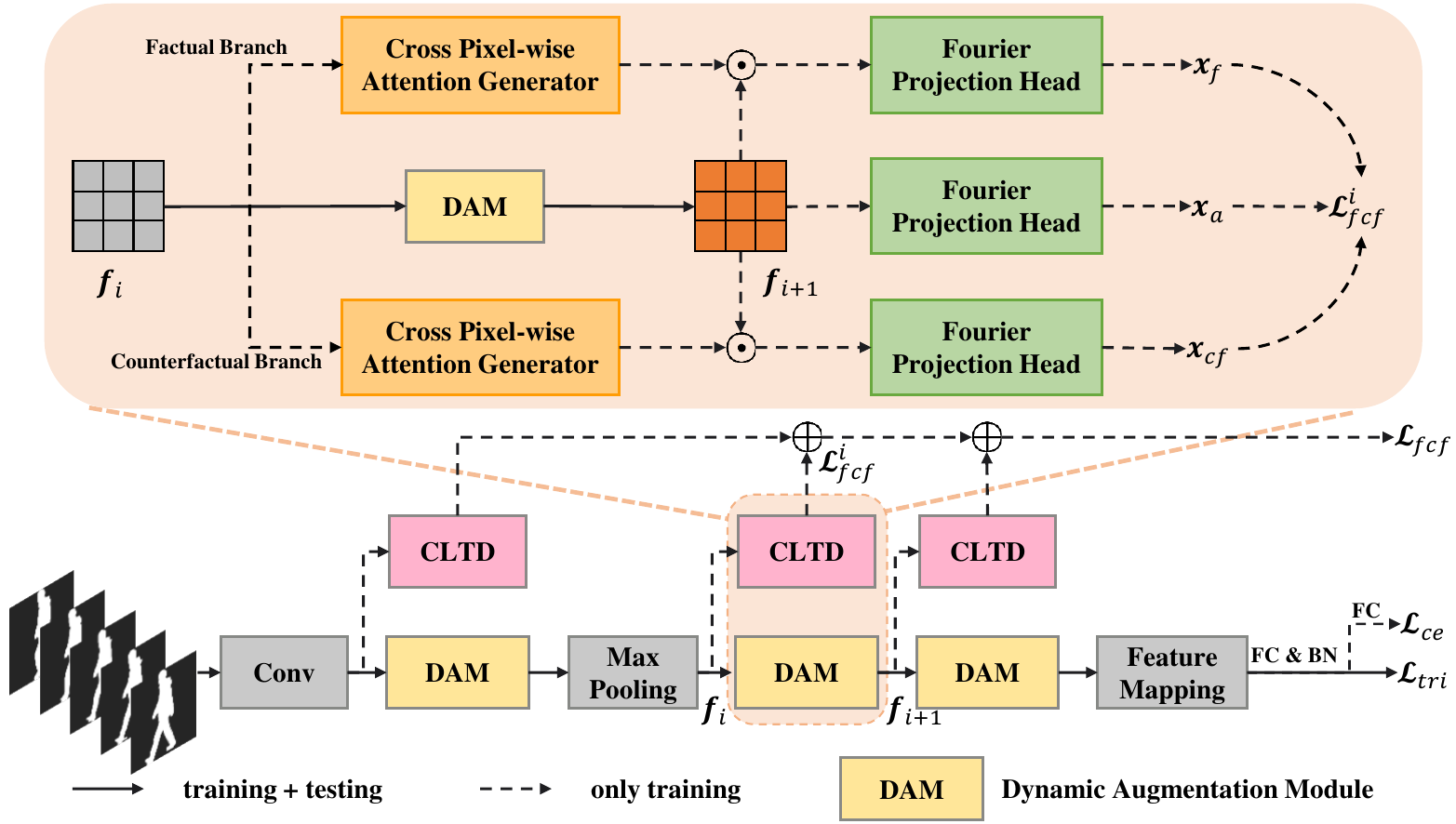}
    \caption{\textbf{Overview of our approach.} We illustrate \name{} using DyGait~\cite{wang2023dygait} as a backbone. In \cref{sec: exp}, we will show its versatility to various gait recognition models. Multiple \name{}s are used along the backbone of DyGait. Each \name{} consists of two branches: the Factual Branch and the Counterfactual Branch. These branches aim to generate the factual feature $\boldsymbol{x}_f$ and the counterfactual feature $\boldsymbol{x}_{cf}$, respectively. Notably, \name{}s are only used for training and excluded during testing.}
    \label{fig: architectural}
\end{figure}

\subsection{Architecture}
Based on the analysis of causality, we introduce \name{} to enhance the learning of discriminative features by eliminating the influence of confounders, while preserving ID-intrinsic information. The overall architecture of our approach is illustrated in~\cref{fig: architectural}. For clarity, we adopt DyGait~\cite{wang2023dygait} as the backbone to construct our approach, and the versatility of \name{} will be demonstrated in~\cref{sec: ablation}. DyGait primarily consists of several fundamental DAM blocks. Given the wide adoption of multi-scale feature representation~\cite{hou2021bicnet,huang2021context,zhao2023movement,wang2023hierarchical} in various computer vision tasks, it is reasonable to infer that confounding information is distributed across multiple feature scales. Therefore, we propose incorporating \name{} at multiple stages of the network, not solely in the final stage, to effectively and comprehensively eliminate the effect of confounders. Specifically, we integrate a \name{} for each DAM to learn more discriminative feature representations across multiple feature grains. Each \name{} is responsible for eliminating the impact of confounders in triple domains by generating factual and counterfactual features simultaneously, supervised with a Factual and Counterfactual Loss. This ensures the gait recognition model remains free from confounders throughout the backbone. The experimental results in~\cref{sec: ablation} will demonstrate the effectiveness of this approach. It is crucial to clarify that \name{} is only employed during training, which incurs no additional computational cost or memory consumption during inference.

\subsection{\name{}}
The proposed \name{} is a plug-and-play module designed to eliminate the impact of confounders in the spatial, temporal, and spectral domains. It consists of two primary components: the Cross Pixel-wise Attention Generator and the Fourier Projection Head.

\noindent \textbf{Cross Pixel-wise Attention Generator.}
The Cross Pixel-wise Attention Generator, inspired by CCNet~\cite{huang2019ccnet}, is introduced to establish high-quality factual and counterfactual attention. This is achieved by employing a separation operation in time, height, and width dimensions. This operation calculates correlations only between pixels that cross a specific pixel, effectively reducing computational complexity in time and space.

\begin{figwindow}[0,r,{
    \includegraphics[width=0.5\linewidth]{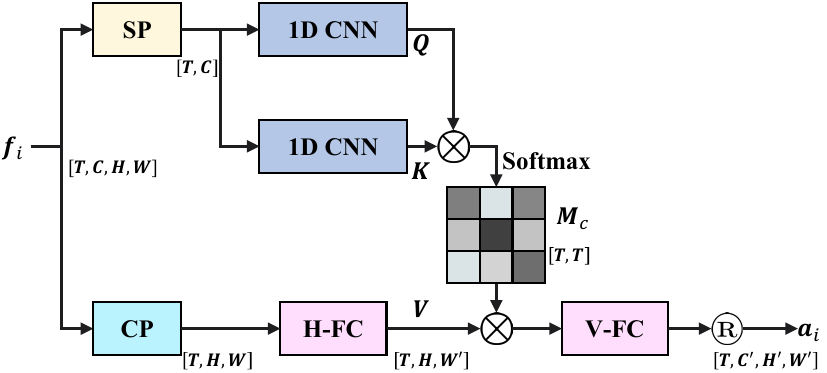}},
    {The detailed structure of CPAG.}
    \label{fig: CPAG}]
    The detailed structure of CPAG depicted in~\cref{fig: CPAG}, involves several key steps. Given the input feature map $\boldsymbol{f}_i \in \mathbb{R}^{T\times C\times H\times W}$, CPAG starts with a Spatial Pooling operation and two 1D CNNs applied to $\boldsymbol{f}_i$, generating distinct feature maps $\boldsymbol{Q}$ and $\boldsymbol{K}$, where ${\boldsymbol{Q}, \boldsymbol{K}} \in \mathbb{R}^{T\times C}$. Subsequently, the temporal correlation matrix $\boldsymbol{M}_c \in \mathbb{R}^{T\times T}$ is generated through a matrix multiplication operation and a softmax layer. The process is formulated as:
\end{figwindow}
\begin{equation}
    \begin{split}
        \boldsymbol{Q} &= \text{Conv1D}\left(\text{SP}(\boldsymbol{f}_i)\right), \\
        \boldsymbol{K} &= \text{Conv1D}\left(\text{SP}(\boldsymbol{f}_i)\right), \\
        \boldsymbol{M}_c &= \text{Softmax}\left(\boldsymbol{Q} \boldsymbol{K}^{\mathsf{T}} / \gamma \right),
    \end{split}
\end{equation}
where $\text{SP}(\cdot)$ denotes the Spatial Pooling operation, and $\gamma$ is a learnable scaling parameter to balance the learning magnitude of the key-query dot product. Simultaneously, CPAG performs an average pooling operation on $\boldsymbol{f}_i$ along the channel dimension. It is then horizontally mapped into $\boldsymbol{V} \in \mathbb{R}^{T\times H\times W'}$ via the Horizontal separate FC layer (H-FC) $\boldsymbol{W}_H \in \mathbb{R}^{H\times W\times W'}$, where $W'$ denotes the width of the \textit{i}-th DAM's output $\boldsymbol{f}_{i+1}$. After obtaining the temporal correlation matrix $\boldsymbol{M}_c$ and feature map $\boldsymbol{V}$, CPAG utilizes matrix multiplication, Vertical separate FC layer (V-FC), and channel repeat operations to produce the output $\boldsymbol{a}_i$, denoted as:
\begin{equation}
    \boldsymbol{a}_i = \text{Repeat}\left(\left(\boldsymbol{M}_c \boldsymbol{V}\right)^{\mathsf{T}} \boldsymbol{W}_V \right),
\end{equation}
where $\boldsymbol{a}_i \in \mathbb{R}^{T\times C'\times H' \times W'}$, and $C'$ and $H'$ refer to the channel and height of $\boldsymbol{f}_{i+1}$, respectively. $\boldsymbol{W}_V \in \mathbb{R}^{W'\times H\times H'}$ represents the V-FC operation like H-FC, and $\text{Repeat}(\cdot)$ represents the channel repeat operation.

In naive self-attention~\cite{vaswani2017attention,dosovitskiy2020image}, the computational complexity of the matrix multiplication is quadratic to the spatial-temporal resolution of the input, i.e., $\mathcal{O}\left(2t^2h^2w^2c\right)$ for a feature map with $t\times c\times h\times w$ pixels. In contrast, our approach, which involves the separation operations of time, height, and width, significantly reduces the computational complexity to $\mathcal{O}\left(t^2(c+hw)\right)$. Therefore, CPAG dramatically lowers the computational cost compared to Self-Attention. Furthermore, horizontal and vertical decoupling enables CPAG to aggregate pixels at different locations along the horizontal and vertical directions, facilitating enhanced spatial interaction at the pixel level. This, in turn, generates high-quality distributions of factual and counterfactual attention.

\noindent \textbf{Fourier Projection Head.}
The distribution of information in the spatial domain is often scattered, posing challenges for effective extraction. In contrast, the Fourier spectrum typically concentrates most of the energy in the low-frequency region, making it more amenable to information extraction. Motivated by this observation, we introduce the Fourier Projection Head to transform the spatial feature map into the Fourier spectral domain through FFT. The subsequent step involves a Low-frequency Selector (LFS) to reduce dimensionality while preserving essential information.

\begin{figure}[tb]
    \centering
    \includegraphics[width=0.7\linewidth]{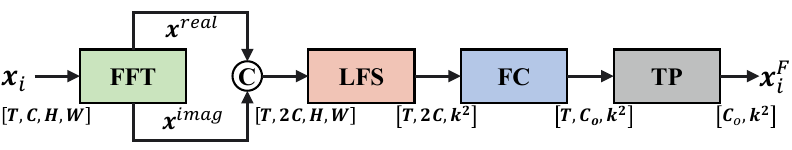}
    \caption{Detailes of FPH. The symbol \textcircled{c} stands for the concatenating operation.}
    \label{fig: FPH}
\end{figure}

As depicted in~\cref{fig: FPH}, FPH takes $\boldsymbol{x}_i = \boldsymbol{a}_i \odot \boldsymbol{f}_{i+1} \in \mathbb{R}^{T\times C\times H\times W}$ as input and first employs FFT to convert $\boldsymbol{x}_i$ into the Fourier spectrum. Subsequently, the LFS is applied to the Fourier spectrum with a window of $k\times k$ to reduce dimensionality. The information is then projected into a vector $\boldsymbol{x}^F_i$ through an FC layer $\boldsymbol{W}_P \in \mathbb{R}^{2C\times 2C_o}$ and a Temporal Pooling (TP) operation, where $C_o$ is the output dimension. The overall FPH process can be formulated as:
\begin{equation}
    \begin{split}
         &\boldsymbol{x}^{real}, \boldsymbol{x}^{imag} = \text{FFT}(\boldsymbol{x}_i),\\
         &\boldsymbol{x}^k = \text{LFS}_{k\times k}(\boldsymbol{x}^{real} \text{\textcircled{c}} \boldsymbol{x}^{imag}),\\
         &\boldsymbol{x}^F_i = \text{TP}(\boldsymbol{x}^k \boldsymbol{W}_P),
    \end{split}
\end{equation}
where $\boldsymbol{x}^F_i \in \mathbb{R}^{C_o\times k^2}$, \textcircled{c} denotes concatenating the real and imaginary components, and LFS$_{k\times k}$ represents the low-frequency selector with a window of $k\times k$, defined as:
\begin{equation}
\text{LFS}_{k\times k}(x) = \text{Sigmoid}(x[...,\frac{h-k}{2}:\frac{h-k}{2}+k, \frac{w-k}{2}:\frac{w-k}{2}+k]).
\end{equation}

\subsection{Loss Function}
We introduce the Factual and Counterfactual Loss as a supervision mechanism for each \name{} in our approach. For the \textit{i}-th \name{}, we obtain the factual feature $\boldsymbol{x}_f$, and counterfactual feature $\boldsymbol{x}_{cf}$. They are used along with the anchor feature $\boldsymbol{x}_a$, the output of FPH taking $\boldsymbol{f}_{i+1}$ as input, to formulate contrastive learning. Here we have omitted the symbol \textit{i} for simplicity. Specifically, we employ InfoNCE~\cite{gutmann2010noise,oord2018representation} to minimize the gap between entangled and ID-intrinsic features while increasing the gap between entangled and confounder features, ensuring consistent semantic representations of ID-intrinsic clues across sequences from the same subject. This is formulated as:
\begin{equation}
    \mathcal{L}_{NCE}^{fcf} = -\sum_{\boldsymbol{x}^+ \in \boldsymbol{S}_f}\log{\frac{sim(\boldsymbol{x}_a, \boldsymbol{x}^+)}{sim(\boldsymbol{x}_a, \boldsymbol{x}^+) + \sum_{\boldsymbol{x} \in \boldsymbol{S}_{cf}}sim(\boldsymbol{x}_a, \boldsymbol{x})}},
\end{equation}
where $sim(\boldsymbol{a},\boldsymbol{b}) = \frac{\boldsymbol{a} \cdot \boldsymbol{b}}{\left\Vert \boldsymbol{a} \right\Vert \left\Vert \boldsymbol{b} \right\Vert}$ represents the similarity between two feature vectors $\boldsymbol{a}$ and $\boldsymbol{b}$. $\boldsymbol{S}_f$ and $\boldsymbol{S}_{cf}$ denote the sets of factual features and counterfactual features, respectively, from the subject that $\boldsymbol{x}_a$ belongs to. Additionally, inspired by the work of Tang \etal~\cite{tang2020unbiased} in alleviating context bias through the total direct effect (TDE) in causal inference for unbiased scene graph generation, we eliminate the impact of confounders by maximizing TDE and the factual probability. This is defined as:
\begin{equation}
    \begin{split}
    &\boldsymbol{Y}_f = \textbf{P}(\boldsymbol{Y} \vert \boldsymbol{X} = \boldsymbol{x}_f) = \boldsymbol{W}_c \boldsymbol{x}_f, \\
    &\boldsymbol{Y}_{cf} = \textbf{P}(\boldsymbol{Y} \vert do(\boldsymbol{X} = \boldsymbol{x}_{cf})) =\boldsymbol{W}_c \boldsymbol{x}_{cf}, \\
    & \boldsymbol{TDE} = \boldsymbol{Y}_f - \boldsymbol{Y}_{cf}, \\
    &\mathcal{L}_{ce}^{fcf} = \mathcal{L}_{ce}(\boldsymbol{TDE}, y) + \mathcal{L}_{ce}(\boldsymbol{Y}_f, y),
    \end{split}
\end{equation}
where $\boldsymbol{W}_c$ and $y$ denote the ID classifier and the ground truth, respectively. The minimization of $\mathcal{L}_{ce}(\boldsymbol{Y}_f, y)$ and $\mathcal{L}_{ce}(\boldsymbol{TDE}, y)$ enforces the factual feature $\boldsymbol{x}_f$ comprising all ID-intrinsic information, and the counterfactual feature $\boldsymbol{x}_{cf}$ only comprising non-ID clues, respectively. Thus, the influence of confounders can be eliminated by the TDE operation. The Factual and Counterfactual Loss of the \textit{i}-th \name{} is expressed as:
\begin{equation}
    \mathcal{L}_{fcf}^i = \mathcal{L}_{NCE}^{fcf} + \mathcal{L}_{ce}^{fcf}.
\end{equation}

The overall loss function is the combination of the triplet loss $\mathcal{L}_{tri}$~\cite{hermans2017defense}, cross-entropy loss $\mathcal{L}_{ce}$, and multi-stage Factual and Counterfactual Loss $\mathcal{L}_{fcf}$:
\begin{equation}
    \mathcal{L} = \mathcal{L}_{tri} + \mathcal{L}_{ce} + \sum_{i=1}^3 \lambda_i \cdot \mathcal{L}_{fcf}^i,
    \label{eq: alllose}
\end{equation}
where $\lambda_i$ is a hyper-parameter to balance the differences among features at various stages.

\section{Experiments}
\label{sec: exp}
We conduct extensive experiments on four popular datasets, \ie, OU-MVLP~\cite{takemura2018multi}, CASIA-B~\cite{yu2006framework}, GREW~\cite{zhu2021gait}, and Gait3D~\cite{zheng2022gait}, to demonstrate the effectiveness of the proposed approach.

\subsection{Datasets}
OU-MVLP~\cite{takemura2018multi} encompasses a significant number of indoor gait samples, totaling 10307 subjects. Each subject is represented by 14 view angles, evenly distributed between 0$^\circ$ to 90$^\circ$ and 180$^\circ$ to 270$^\circ$, with each angle containing two sequences (Seq\#00-01). According to the protocol outlined in~\cite{takemura2018multi,lin2021gait}, a total of 5153 subjects are employed as the training set, while the remaining 5154 are allocated for testing. CASIA-B~\cite{yu2006framework} comprises 124 subjects, each captured from 11 distinct view angles evenly spread across a range from 0$^\circ$ to 180$^\circ$. Within each view angle, there are three different walking conditions: normal walking (NM\#1-6), walking with bags (BG\#1-2), and walking with coats (CL\#1-2). Following the established protocol detailed in~\cite{chao2019gaitset}, the initial 74 sequences are used for training. GREW~\cite{zhu2021gait} is a large-scale wild dataset containing a vast array of 26345 subjects and 128671 sequences. GREW is divided into three distinct parts: the training set (20000 subjects), the validation set (345 subjects), and the test set (6000 subjects). Gait3D~\cite{zheng2022gait} is a wild dataset comprising 4000 subjects and 25309 sequences. Following its protocol, 3000 subjects are designated for the training set, while the remaining subjects are allocated for the test set.

\subsection{Implementation Details}
Our approach is implemented using the PyTorch framework~\cite{paszke2019pytorch}, and all experiments are conducted on NVIDIA GeForce RTX3090 GPUs. For all experiments, the default input silhouette size is $64\times 44$.

\noindent\textbf{Baseline Models.} To demonstrate the effectiveness of \name{}, we employ DyGait~\cite{wang2023dygait} as a backbone for comparison with some advanced methods. Additionally, to showcase the versatility of \name{}, we also use GaitSet~\cite{chao2019gaitset}, GaitPart~\cite{fan2020gaitpart}, GaitGL~\cite{lin2021gait}, GaitGCI~\cite{dou2023gaitgci}, and GaitBase~\cite{fan2023opengait} as baselines, which illustrates the improvements \name{} achieves across various frameworks on diverse datasets.

\begin{table}[tb]
    \caption{Comparison results of Rank-1 (\%) on OU-MVLP, excluding identical-view cases. The best result is indicated in \textbf{bold}. The results of the last 6 lines are obtained by removing invalid probe sequences.}
    \centering
    \resizebox{\linewidth}{!}{
    \begin{tabular}{c|c|cccccccccccccc|c}
        \toprule
        \multirow{2}{*}{Method} & \multirow{2}{*}{Venue} & \multicolumn{14}{c|}{Probe View} & \multirow{2}{*}{Mean} \\ \cline{3-16}
        & & $0^{\circ}$ & $15^{\circ}$ & $30^{\circ}$ & $45^{\circ}$ & $60^{\circ}$ & $75^{\circ}$ & $90^{\circ}$ & $180^{\circ}$ & $195^{\circ}$ & $210^{\circ}$ & $225^{\circ}$ & $240^{\circ}$ & $255^{\circ}$ & $270^{\circ}$ & \\ \hline \hline
        GaitSet~\cite{chao2019gaitset} & AAAI19 & 70.3 & 87.9 & 90.0 & 90.1 & 88.0 & 88.7 & 87.7 & 81.8 & 86.5 & 89.0 & 89.2 & 87.2 & 87.6 & 86.2 & 87.1 \\
        GaitPart~\cite{fan2020gaitpart} & CVPR20 & 82.6 & 88.9 & 90.8 & 91.0 & 89.7 & 89.9 & 89.5 & 85.2 & 88.1 & 90.0 & 90.1 & 89.0 & 89.1 & 88.2 & 88.7 \\
        GLN~\cite{hou2020gait} & ECCV20 & 83.8 & 90.0 & 91.0 & 91.2 & 90.3 & 90.0 & 89.4 & 85.3 & 89.1 & 90.5 & 90.6 & 89.6 & 89.3 & 88.5 & 89.2 \\
        GaitGL~\cite{lin2021gait} & ICCV21 & 84.9 & 90.2 & 91.1 & 91.5 & 91.1 & 90.8 & 90.3 & 88.5 & 88.6 & 90.3 & 90.4 & 89.6 & 89.5 & 88.8 & 89.7 \\
        CSTL~\cite{huang2021context} & ICCV21 & 87.1 & 91.0 & 91.5 & 91.8 & 90.6 & 90.8 & 90.6 & 89.4 & 90.2 & 90.5 & 90.7 & 89.8 & 90.0 & 89.4 & 90.2 \\
        3DLocal~\cite{huang20213d} & ICCV21 & 86.1 & 91.2 & 92.6 & 92.9 & 92.2 & 91.3 & 91.1 & 86.9 & 90.8 & \textbf{92.2} & 92.3 & 91.3 & 91.1 & 90.2 & 90.9 \\
        MetaGait~\cite{dou2022metagait} & ECCV22 & 88.2 & 92.3 & \textbf{93.0} & \textbf{93.5} & 93.1 & \textbf{92.7} & \textbf{92.6} & 89.3 & 91.2 & 92.0 & \textbf{92.6} & 92.3 & \textbf{91.9} & 91.1 & 91.9 \\
        DANet~\cite{ma2023dynamic} & CVPR23 & 87.7 & 91.3 & 91.6 & 91.8 & 91.7 & 91.4 & 91.1 & 90.4 & 90.3 & 90.7 & 90.9 & 90.5 & 90.3 & 89.9 & 90.7 \\ 
        GaitBase~\cite{fan2023opengait} & CVPR23 & - & - & - & - & - & - & - & - & - & - & - & - & - & - & 90.8 \\ 
        GaitGCI~\cite{dou2023gaitgci} & CVPR23 & 91.2 & 92.3 & 92.6 & 92.7 & 93.0 & 92.3 & 92.1 & 92.0 & \textbf{91.8} & 91.9 & \textbf{92.6} & 92.3 & 91.4 & 91.6 & 92.1 \\ 
         \rblue \textbf{Ours} & - & \textbf{91.6} & \textbf{92.5} & 92.7 & 92.6 & \textbf{93.2} & 92.4 & 92.4 & \textbf{92.5} & \textbf{91.8} & \textbf{92.2} & 91.9 & \textbf{92.5} & \textbf{91.9} & \textbf{91.8} & \textbf{92.3} \\ \hline
        GaitSet~\cite{chao2019gaitset} & AAAI19 & 84.5 & 93.3 & 96.7 & 96.6 & 93.5 & 95.3 & 94.2 & 87.0 & 92.5 & 96.0 & 96.0 & 93.0 & 94.3 & 92.7 & 93.3  \\
        GaitPart~\cite{fan2020gaitpart} & CVPR20 & 88.0 & 94.7 & 97.7 & 97.6 & 95.5 & 96.6 & 96.2 & 90.6 & 94.2 & 97.2 & 97.1 & 95.1 & 96.0 & 95.0 & 95.1 \\
        GLN \cite{hou2020gait} & ECCV20 & 89.3 & 95.8 & 97.9 & 97.8 & 96.0 & 96.7 & 96.1 & 90.7 & 95.3 & 97.7 & 97.5 & 95.7 & 96.2 & 95.3 & 95.6 \\
        GaitGL~\cite{lin2021gait} & ICCV21 & 90.5 & 96.1 & 98.0 & 98.1 & 97.0 & 97.6 & 97.1 & 94.2 & 94.9 & 97.4 & 97.4 & 95.7 & 96.5 & 95.7 & 96.2 \\
        DyGait~\cite{wang2023dygait} & ICCV23 & 96.2 & 98.2 & 99.1 & 99.0 & 98.6 & 99.0 & 98.8 & 97.9 & 97.6 & 98.8 & 98.6 & 98.1 & 98.3 & 98.2 & 98.3 \\
        \rblue \textbf{Ours} & - & \textbf{97.0} & \textbf{98.3} & \textbf{99.5} & \textbf{99.2} & \textbf{98.9} & \textbf{99.2} & \textbf{99.1} & \textbf{98.1} & \textbf{98.0} & \textbf{99.1} & \textbf{98.8} & \textbf{98.5} & \textbf{98.7} & \textbf{98.3} & \textbf{98.6} \\
        \bottomrule
    \end{tabular}}
    \label{tab: OU-MVLP}
\end{table}
\noindent\textbf{Training Strategies.} The experimental settings align with the established training strategies of baseline models, including the choice of the optimizer, scheduler, iteration count, and batch size. This way ensures a fair and consistent evaluation framework across all models. More training strategies are provided in the supplementary material.

\noindent\textbf{Hyper Parameters.} To adapt our approach to the unique characteristics of each dataset, we tailor certain parameters accordingly. Specifically, for CASIA-B, we set the output dimension of FPH, denoted as $C_o$ to 128, and the weights ($\lambda_1, \lambda_2, \lambda_3$) in~\cref{eq: alllose} to (0.05, 0.10, 0.15), respectively. For OU-MVLP, GREW, and Gait3D, we increase $C_o$ and the weights ($\lambda_1, \lambda_2, \lambda_3$) to 256 and (0.1, 0.2, 0.3), respectively. Additionally, for all datasets, we set the number of \name{}s to 3, and the window size $k\times k$ of LFS in FPH to $7\times7$.

\subsection{Comparison with State-of-the-art Methods}
In this section, we compare our approach with several methods~\cite{chai2022lagrange,chao2019gaitset,fan2020gaitpart,hou2020gait,lin2021gait,huang2021context,huang20213d,dou2022metagait,fan2023opengait,dou2023gaitgci,zheng2022mhop,wang2023hierarchical,ma2023dynamic,wang2023dygait,zheng2022gait,wang2023causal}. More experimental results are provided in the supplementary material.

\begin{tabwindow}[0,r,{
\resizebox{0.5\linewidth}{!}{
    \begin{tabular}{c|c|cccc}
        \toprule
        Method & Venue & NM & BG & CL & Mean \\ \hline\hline
        GaitSet~\cite{chao2019gaitset} & AAAI19 & 95.0 & 87.2 & 70.4 & 84.2 \\
        GaitPart~\cite{fan2020gaitpart} & CVPR20 & 96.2 & 91.5 & 78.7 & 88.8 \\
        GLN~\cite{hou2020gait} & ECCV20 & 96.9 & 94.0 & 77.5 & 89.5 \\
        GaitGL~\cite{lin2021gait} & ICCV21 & 97.4 & 94.5 & 83.6 & 91.8 \\
        3DLocal~\cite{huang20213d} & ICCV21 & 97.5 & 94.4 & 83.7 & 91.8 \\
        CSTL~\cite{huang2021context} & ICCV21 & 97.8 & 93.6 & 84.2 & 91.9 \\
        LagrangeGait~\cite{chai2022lagrange} & CVPR22 & 96.9 & 93.5 & 86.5 & 92.3 \\ 
        MetaGait~\cite{dou2022metagait} & ECCV22 & 98.1 & 95.2 & 86.9 & 93.4 \\ 
        DANet~\cite{ma2023dynamic} & CVPR23 & 98.0 & 95.9 & \textbf{89.9} & 94.6 \\
        GaitBase~\cite{fan2023opengait} & CVPR23 & 97.6 & 94.0 & 77.4 & 89.6 \\
        GaitGCI~\cite{dou2023gaitgci} & CVPR23 & 98.4 & \textbf{96.6} & 88.5 & 94.5 \\
        HSTL~\cite{wang2023hierarchical} & ICCV23 & 98.1 & 95.9 & 88.9 & 94.3 \\ 
        DyGait~\cite{wang2023dygait} & ICCV23 & 98.4 & 96.2 & 87.8 & 94.1 \\ \hline
        \rblue \textbf{Ours} & - & \textbf{98.6} & 96.4 & 89.3 & \textbf{94.8} \\
        \bottomrule
    \end{tabular}
    }},
    {Comparison results of Rank-1 (\%) with SOTA methods on CASIA-B, excluding identical-view case.}
    \label{tab: CASIA-B}]
    \noindent\textbf{Evaluation on OU-MVLP.} As indicated  in~\cref{tab: OU-MVLP}, we compare \name{} with previous methods on OU-MVLP. Our proposed approach outperforms the existing methods across most view angles (9 out of 14), which demonstrates its effectiveness. Specifically, it surpasses GaitGCI~\cite{dou2023gaitgci} and DyGait~\cite{wang2023dygait} by 0.2\% and 0.3\%, respectively, establishing SOTA performance and affirming its effectiveness.

    \noindent\textbf{Evaluation on CASIA-B.} As illustrated in~\cref{tab: CASIA-B}, our approach undergoes a comparative analysis with published methods on three walking conditions under identical evaluation settings. Observing the table, our approach achieves an average accuracy of 94.8\%, which is 0.2\% and 0.3\% higher than DANet~\cite{ma2023dynamic} and GaitGCI~\cite{dou2023gaitgci}, respectively, establishing itself as the SOTA. Furthermore, compared to DyGait~\cite{wang2023dygait}, the integration of \name{} yields remarkable performance improvement, with an average improvement of 0.7\%, particularly evident under the CL condition, where it increases by 1.5\%. This underscores the strong discriminative feature learning capability of our approach, attributed to the effective elimination of confounders.
\end{tabwindow}

It is important to note that the performance improvements on OU-MVLP and CASIA-B may not appear particularly significant, with Rank-1 accuracy improvement of 0.3\% and 0.7\% respectively. This is likely due to the relative simplicity of these datasets, where performance tends to reach saturation. However, the benefits of \name{} will become more pronounced when applied to more challenging datasets, as demonstrated in the following experiments.

\begin{table}[tb]
    \caption{Comparison of Rank-1 and Rank-5 (\%) on GREW and Giat3D.}
    \centering
    \resizebox{0.7\linewidth}{!}{
    \begin{tabular}{c|c|cc|cc}
        \toprule
        \multirow{2}{*}{Method} & \multirow{2}{*}{Venue} &  \multicolumn{2}{c|}{GREW} & \multicolumn{2}{c}{Gait3D} \\ \cline{3-6}
        & & Rank-1 & Rank-5 & Rank-1 & Rank-5 \\ \hline \hline
        GaiSet~\cite{chao2019gaitset}   & AAAI19 & 46.3 & 63.6 & 36.7 & 58.3 \\
        GaitPart~\cite{fan2020gaitpart} & CVPR20 & 44.0 & 60.7 & 28.2 & 47.6 \\
        GaitGL~\cite{lin2021gait}       & ICCV21 & 47.3 & 63.6 & 29.7 & 48.5 \\
        SMPLGait~\cite{zheng2022gait}   & CVPR22 & -    & -    & 46.3 & 64.5 \\ 
        MTSGait~\cite{zheng2022mhop}    & ACM MM22 & 55.3 & 71.3 & 48.7 & 67.1 \\ 
        DANet~\cite{ma2023dynamic}      & CVPR23 & -    & -    & 48.0 & 69.7 \\
        GaitBase~\cite{fan2023opengait}   & CVPR23 & 60.1 & -    & 64.6 & -  \\
        GaitGCI~\cite{dou2023gaitgci} & CVPR23 & 68.5 & 80.8 & 50.3 & 68.5 \\
        HSTL~\cite{wang2023hierarchical}   & ICCV23 & 62.7 & 76.6   & 61.3 & 76.3  \\
        DyGait~\cite{wang2023dygait}    & ICCV23 & 71.4 & 83.2 & 66.3 & 80.8 \\
        GaitCSV~\cite{wang2023causal}   & ACM MM23 & 64.9 & - & 69.1 & - \\    \hline
        \rblue \textbf{Ours} & - & \textbf{78.0} & \textbf{87.8} & \textbf{69.7} & \textbf{85.2} \\ 
        \bottomrule
    \end{tabular}
    }
    \label{tab: GREW and Gait3D}
\end{table}

\noindent\textbf{Evaluation on GREW and Gait3D.} Validation of the effectiveness of our approach on two wild datasets, namely GREW and Gait3D, is presented in~\cref{tab: GREW and Gait3D}. Compared with DyGait~\cite{wang2023dygait} and GaitCSV~\cite{wang2023causal}, our approach outperforms by 6.6\% and 13.1\% on GREW, and by 3.4\% and 0.6\% on Gait3D, respectively. This highlights the capability of our approach to eliminate the influence of confounders effectively, capture more discriminative features, and contribute to better performance. When considering the results collectively from~\cref{tab: OU-MVLP},~\cref{tab: CASIA-B}, and~\cref{tab: GREW and Gait3D}, it is evident that other methods experience a significant performance drop on wild datasets due to increased confounders. However, our approach still achieves 78.0\% Rank-1 accuracy on GREW, establishing itself as the SOTA, and demonstrating strong robustness across diverse scenarios.

\subsection{Ablation Study}
\label{sec: ablation}
In this section, we conduct ablation experiments on GREW and Gait3D to verify the design of \name{}. 

\begin{table}[tb]
    \caption{Performance improvements (Rank-1 (\%)) when using \name{} with different baselines. Identical-view cases are excluded on CASIA-B and OU-MVLP. The symbol $^\dagger$ represents excluding invalid probe sequences on OUMVLP. $^{\ast}$ indicates GaitGCI without its CIL block.}
    \centering
    \resizebox{0.7\linewidth}{!}{
    \begin{tabular}{l|c|c|c|c|c}
        \toprule
        \multirow{2}{*}{Backbone} &\multirow{2}{*}{Venue} &  \multicolumn{4}{c}{Testing Dataset} \\ \cline{3-6}
        & & CASIA-B & OU-MVLP & GREW & Gait3D\\ \hline \hline
        GaiSet~\cite{chao2019gaitset} & AAAI19 & 84.2 & 87.1 & 46.3 & 36.7 \\
        \rblue GaiSet w/\name{} & - & \textbf{88.1}\color{red}{$^{\uparrow 2.2}$} & \textbf{89.3}\color{red}{$^{\uparrow 2.2}$} & \textbf{51.8}\color{red}{$^{\uparrow 5.5}$} & \textbf{40.8}\color{red}{$^{\uparrow 4.1}$} \\ \hline
        GaitPart~\cite{fan2020gaitpart} & CVPR20 & 88.8 & 88.7 & 44.0 & 28.2 \\
        \rblue GaitPart w/\name{} & - & \textbf{91.5}\color{red}{$^{\uparrow 2.7}$} & \textbf{90.1}\color{red}{$^{\uparrow 1.4}$} & \textbf{51.1}\color{red}{$^{\uparrow 7.1}$} & \textbf{31.4}\color{red}{$^{\uparrow 3.2}$} \\ \hline
        GaitGL~\cite{lin2021gait} & ICCV21 & 91.8 & 89.7 & 47.3 & 29.7 \\
        \rblue GaitGL w/\name{} & - & \textbf{93.5}\color{red}{$^{\uparrow 1.7}$} & \textbf{90.8}\color{red}{$^{\uparrow 1.1}$} & \textbf{57.4}\color{red}{$^{\uparrow 10.1}$} & \textbf{33.2}\color{red}{$^{\uparrow 3.5}$} \\ \hline
         GaitBase~\cite{fan2023opengait} & CVPR23 & 89.6 & 90.8 & 60.1 & 64.6 \\
         \rblue GaitBase w/\name{} & - & \textbf{92.4}\color{red}{$^{\uparrow 2.8}$} & \textbf{91.3}\color{red}{$^{\uparrow 0.5}$} & \textbf{70.4}\color{red}{$^{\uparrow 10.3}$} & \textbf{71.9}\color{red}{$^{\uparrow 7.3}$} \\ \hline
        GaitGCI~\cite{dou2023gaitgci} & CVPR23 &  93.1 & 92.1 & 68.5 & 50.3 \\
        {\color{gray} GaitGCI$^{\ast}$} & - & {\color{gray} 90.1} & {\color{gray}90.2} & {\color{gray}60.2} & {\color{gray}45.8} \\
        \rblue GaitGCI$^{\ast}$ w/\name{} & - &  \textbf{93.5}\color{red}{$^{\uparrow 3.4}$} & \textbf{92.3}\color{red}{$^{\uparrow 2.1}$} & \textbf{71.3}\color{red}{$^{\uparrow 11.1}$} & \textbf{52.2}\color{red}{$^{\uparrow 6.4}$} \\ \hline
        DyGait$^\dagger$~\cite{wang2023dygait} & ICCV23 & 94.1 & 98.3 & 71.4 & 66.3 \\
         \rblue DyGait$^\dagger$ w/\name{} & - &  \textbf{94.8}\color{red}{$^{\uparrow 0.7}$} & \textbf{98.6}\color{red}{$^{\uparrow 0.3}$} & \textbf{78.0}\color{red}{$^{\uparrow 6.6}$} & \textbf{69.7}\color{red}{$^{\uparrow 3.4}$}  \\
         \bottomrule
    \end{tabular}}
    \label{tab: versatility}
\end{table}

\noindent\textbf{Versatility of \name{}.} In~\cref{tab: versatility}, we investigate the versatility of \name{} with different gait recognition models, including GaitSet~\cite{chao2019gaitset}, GaitPart~\cite{fan2020gaitpart}, GaitGL~\cite{lin2021gait}, GaitBase~\cite{fan2023opengait}, GaitGCI~\cite{dou2023gaitgci}, and DyGait~\cite{wang2023dygait}. From~\cref{tab: versatility}, we can summarize the following valuable findings: (1) When integrated with our approach, all methods exhibit performance improvements on four datasets, indicating the efficacy of \name{}. (2) Particularly noteworthy is the significant performance improvement observed on GREW, with a maximum improvement of \textbf{11.1\%} (from 60.2\% to 71.3\%) when using GaitGCI without its causal module CIL as the backbone.

\begin{tabwindow}[0,r,{
    \resizebox{0.45\linewidth}{!}{
    \begin{tabular}{l|ccc}
        \toprule
        Method &  GREW & Gait3D   \\ \hline \hline
        Baseline & 71.4 & 66.3 \\ 
        Baseline + CPAG & 73.3 & 67.1 \\
        Baseline + CPAG + FPH& 76.0 & 68.6 \\ \hline
        \rblue \textbf{Ours} & \textbf{78.0} & \textbf{69.7} \\
        \bottomrule
    \end{tabular}}
},{Study the effectiveness of proposed modules in \name{} on GREW and Gait3D, including CPAG, FPH and Factual and Counterfactual Loss.}
\label{tab: ablation1}]
\noindent\textbf{Effectiveness of proposed components.} To assess the effectiveness of the proposed components, we conduct ablation experiments on GREW and Gait3D. As shown in~\cref{tab: ablation1}, it can be observed that the integration of proposed modules leads to consistent performance improvements. The CPAG contributes to the improvement of recognition performance, with an average improvement of 1.9\% on GREW compared to the baseline. Additionally, employing FPH and Factual and Counterfactual Loss together yields 5.7\% performance improvement on GREW, underscoring the effectiveness of these components.
\end{tabwindow}

\begin{tabwindow}[0,r,{
    \resizebox{0.45\linewidth}{!}{
    \begin{tabular}{c|ccc|cc}
        \toprule
        Method &1-\textit{st} & 2-\textit{nd} & 3-\textit{rd} & GREW & Gait3D  \\ \hline \hline
        Baseline & & & & 71.4 & 66.3  \\ \hline
        a & $\checkmark$ & & & 73.6 & 67.0  \\
        b & & $\checkmark$ & & 74.2 & 67.5  \\
        c & & & $\checkmark$ & 75.1 & 68.4  \\
        d & $\checkmark$ & $\checkmark$ & & 75.4 & 68.6  \\
        e & $\checkmark$ & & $\checkmark$ & 76.5 & 69.1  \\
        f & & $\checkmark$ & $\checkmark$ & 77.1 & 69.3  \\ 
        \rblue g & $\checkmark$ & $\checkmark$ & $\checkmark$ & \textbf{78.0} & \textbf{69.7} \\ 
        \bottomrule
    \end{tabular}}
},{Impact of multiple stages.}
\label{tab: ablation2}]
\noindent\textbf{Impact of multiple stages.} In~\cref{tab: ablation2}, we analyze the impact of the number and position of \name{}s in terms of accuracy. From the results, we note that: (\textbf{1}) The impact of using \name{} at different positions on performance varies. The deeper layer it is used at, the more significant performance improvement is achieved. This phenomenon indicates the importance of eliminating confounders closer to the output layer. (\textbf{2}) The best performance is achieved when \name{}s are used simultaneously at multiple positions, resulting in a 6.6\%  improvement compared to the baseline on GREW. This highlights the complementary nature of \name{} when integrated at different positions within the network.
\end{tabwindow}

\begin{figure}[tb]
    \centering
    \includegraphics[width=\linewidth]{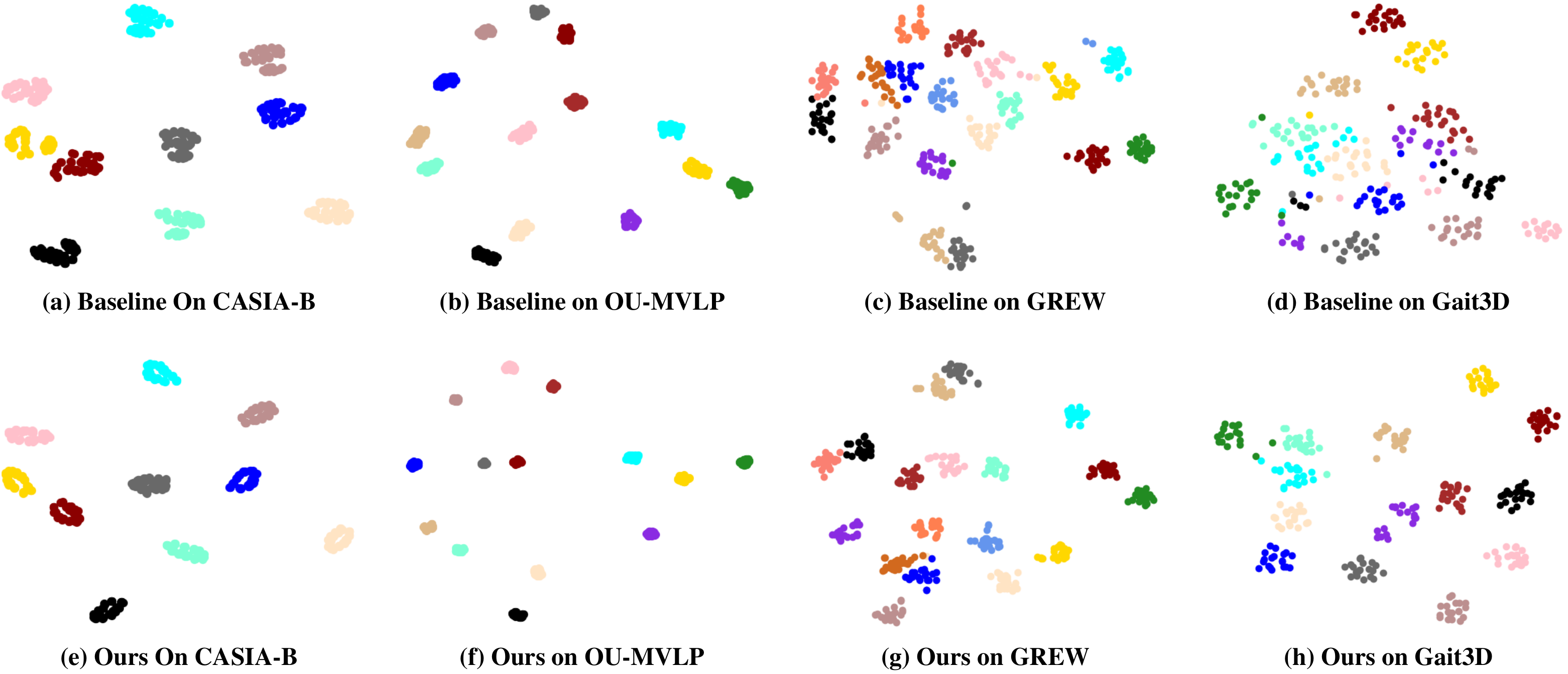}
    \caption{t-SNE visualization examples of feature distributions between the baseline and our approach on CASIA-B, OU-MVLP, GREW, and Gait3D, respectively. Different colors denote distinct identities. Best viewed by color and zooming in.}
    \label{fig: visual}
\end{figure}

\begin{tabwindow}[0,r,{\resizebox{0.4\linewidth}{!}{
    \begin{tabular}{c|cc|cc}
        \toprule
        Method & Cross-Entropy & InfoNCE & GREW & Gait3D  \\ \hline \hline
        Baseline & & &        71.4 & 66.3 \\ \hline
        a & $\checkmark$ & &   76.0 & 68.6  \\
        b & & $\checkmark$ &   77.3 & 69.2 \\
        \rblue
        c & $\checkmark$ & $\checkmark$    & \textbf{78.0} & \textbf{69.7} \\ 
        \bottomrule
    \end{tabular}}},
{Impact of loss function.}
\label{tab: ablation3}]
    \noindent\textbf{Effectiveness of factual cross-entropy loss and InfoNCE loss.} As reported in~\cref{tab: ablation3}, we conduct experiments to investigate the effectiveness of factual cross-entropy loss and InfoNCE loss. We find that both losses contribute to improved recognition performance, and using them together results in even higher performance, demonstrating their complementary properties.
\end{tabwindow}

\begin{tabwindow}[0,r,{\resizebox{0.4\linewidth}{!}{
    \begin{tabular}{c|cc|cc}
        \toprule
        \multirow{2}{*}{$k\times k$} & \multicolumn{2}{c|}{GREW} & \multicolumn{2}{c}{Gait3D}\\ \cline{2-5}
         & Rank-1 & Rank-5 & Rank-1 & Rank-5  \\ \hline \hline
        $3\times 3$ & 76.4 & 85.2 & 66.7 & 81.6 \\
        $5\times 5$ & 77.0 & 86.4 & 68.3 & 83.9 \\
        \rblue
        $7\times 7$ & \textbf{78.0} & \textbf{87.8} & \textbf{69.7} & \textbf{85.2} \\
        $9\times 9$ & 77.2 & 86.6 & 68.9 & 84.7 \\
        $11\times 11$ & 76.3 & 84.9 & 67.4 & 82.5 \\
        \bottomrule
    \end{tabular}}},
{Impact of the window size $k\times k$ in FPH.}
\label{tab: ablation4}]
\noindent\textbf{Impact of the window size $k\times k$ in FPH.} To study the impact of the value $k$, we conduct five experiments detailed in~\cref{tab: ablation4} with $k$ of 3, 5, 7, 9, and 11, respectively. The average accuracy initially increases as $k$ rises, reaching a peak, and then degrades with larger $k$. This trend indicates that very small $k$ may not adequately explore discriminative features, while very large $k$ may distract discriminative features with less discriminative features. Based on the results in~\cref{tab: ablation4}, $k=7$ is chosen for the best performance.
\end{tabwindow}

\subsection{Qualitative results}

\noindent \textbf{Visualization of feature distributions.} We employ t-SNE~\cite{van2008visualizing} to visualize the feature distributions between the baseline and our approach on CASIA-B, OU-MVLP, GREW, and Gait3D. As shown in~\cref{fig: visual}, we observe that the feature distributions produced by our approach are more compact for the same subject compared to the baseline. Consequently, identities become easier to distinguish. These visualizations confirm that our approach effectively eliminates the interference of confounders and extracts the discriminative features that are genuinely relevant for identification. This provides additional support for the effectiveness of our approach.

\begin{figure}[tb]
    \centering
    \includegraphics[width=0.7\linewidth]{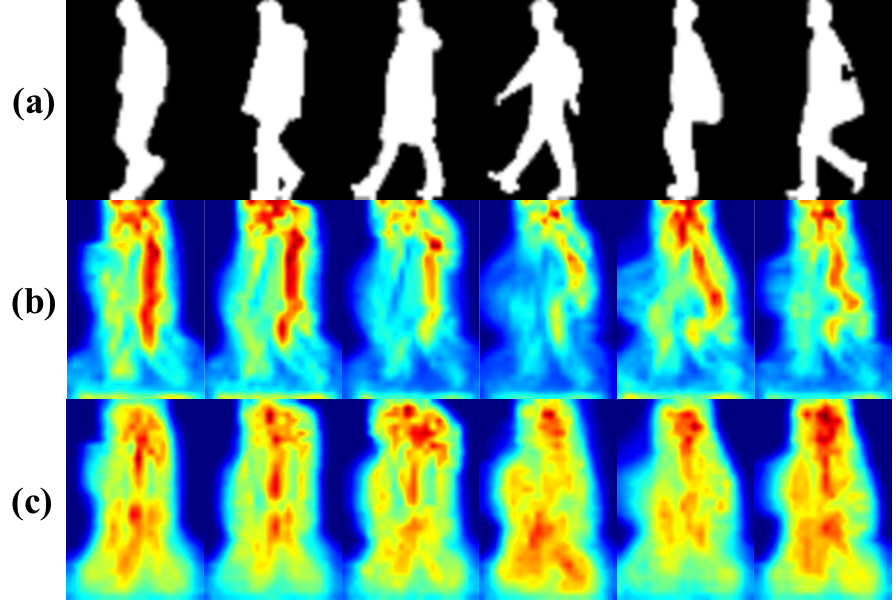}
    \caption{Visualization of heatmaps representing counterfactual and factual features after \name{}. Best viewed in color. (\textbf{a}) Original input. (\textbf{b}) Counterfactual features. (\textbf{c}) Factual features.}
    \label{fig: visual2}
\end{figure}

\noindent \textbf{Visualization of heatmaps.} As illustrated in~\cref{fig: visual2}, we present visualizations of the feature heatmaps processed after \name{}. The objective is to illustrate the superiority of our approach qualitatively. A notable observation is that counterfactual attention tends to focus more on confounders such as clothing and bagging, which are unrelated to identity. In contrast, factual attention concentrates on ID-intrinsic information. This observation demonstrates that \name{} can effectively eliminate the influence of confounders.

\section{Conclusion}
In this work, we present a discriminative feature learning module \name{} based on causality. \name{} effectively eliminates the impact of confounders in spatial, temporal, and spectral domains. Thorough quantitative and qualitative experimental analyses demonstrate the effectiveness and versatility of our approach. Future work will explore the application of causality in other computer vision tasks, such as action recognition, person re-identification, and image restoration.

\noindent \textbf{Limitation.} FPH can improve performance while reducing time costs, but feature selection is fixed. Next, we consider using the attention mechanism for adaptive selection.

\section*{Acknowledgements}
This work was supported by the National Key R\&D Program of China under project 2023YFF0905401.

\bibliographystyle{splncs04}
\bibliography{eccv}
\end{document}